\documentclass[runningheads]{llncs}
\usepackage[T1]{fontenc}
\usepackage{graphicx,verbatim}
\usepackage{cite}
\usepackage{array}
\usepackage{booktabs}
\usepackage{multirow}
\usepackage{adjustbox}
\usepackage{amsmath,amssymb,amsfonts}
\usepackage{algorithmic}
\usepackage{graphicx}
\usepackage{textcomp}
\usepackage{xcolor}
\begin{document}
\title{A Prototype-Guided Coarse Annotations Refining Approach for Whole Slide Images}
%
% \begin{comment}  %% Removed for anonymized MICCAI 2025 submission
\author{Bingjian Yao \and
Weiping Lin \and
Yan He \and
Zheng Wang \and 
Liangsheng Wang}
\authorrunning{B.Yao et al.}
\institute{Department of Computer Science and Technology, School of Informatics, Xiamen University}

% \end{comment}

% \author{Anonymized Authors}  %% Added for anonymized MICCAI 2025 submission
% \authorrunning{Anonymized Author et al.}
% \institute{Anonymized Affiliations \\
%     \email{email@anonymized.com}}

\maketitle              % typeset the header of the contribution
\begin{abstract}
% 介绍背景
% Artificial intelligence-based computational pathology has significantly advanced the efficiency of cancer diagnosis. 
%In computational pathology, pathological images are typically scanned as whole slide images (WSIs) at extremely high resolutions, where 
The fine-grained annotations in whole slide images (WSIs) show the boundaries of various pathological regions. However, generating such detailed annotation is often costly, whereas the coarse annotations are relatively simpler to produce. %Consequently, refining coarse annotations represents a promising and practical approach for improving diagnostic workflows.
% 现有方法
Existing methods for refining coarse annotations often rely on extensive training samples or clean datasets, and fail to capture both intra-slide and inter-slide latent sematic patterns, limiting their precision. In this paper, we propose a prototype-guided approach. %This method requires only a small amount of training data and does not depend on highly accurate initial annotations, offering a more efficient and practical solution. 
Specifically, we introduce a local-to-global approach to construct non-redundant representative prototypes by jointly modeling intra-slide local semantics and inter-slide contextual relationships. Then a prototype-guided pseudo-labeling module is proposed for refining coarse annotations. Finally, we employ dynamic data sampling and re-finetuning strategy to train a patch classifier.
% To address this, we propose a prototype-guided approach. Specifically, we propose a local-to-global prototyping method to construct a dictionary that captures diverse representative pathological patterns. Pseudo-labels for each patch are then generated by querying this dictionary. 
% Finally, a classifier is trained to predict for each patch, which can subsequently be transformed into precise fine-grained annotations. 
Extensive experiments on three publicly available WSI datasets, covering lymph, liver, and colorectal cancers, demonstrate that our method significantly outperforms existing state-of-the-art (SOTA) methods. The code will be available.

\keywords{Coarse annotations refining  \and Prototype learning \and Whole slide image.}
% Authors must provide keywords and are not allowed to remove this Keyword section.
\end{abstract}
\section{Introduction}
% 细粒度标注的困难
Pathological images are considered the gold standard for cancer diagnosis \cite{yao2020pathological}, typically scanned as WSIs with gigapixel resolutions (e.g., $100,000\times100,000$ pixels at $40\times$ magnification). 
Although deep learning-based methods \cite{Chan_2019_ICCV,chen2017dcan,xu2017gland,graham2019mild} have demonstrated excellent performance across various pathological image analysis tasks, their success heavily depends on sufficient fine-grained annotations. However, generating such annotations is both labor-intensive and time-consuming \cite{dimitriou2019deep} and suffers from inter- and intra-observer variability \cite{longacre2006interobserver,foss2012inter,van2013interobserver}, particularly when delineating morphologically ambiguous pathological subtypes.
% Even for experienced pathologist, such annotation for a WSI still takes several hours. 

% 粗标注的 # 这一段是否有必要加一个精标注和粗标注的图
Therefore, coarse annotations emerge as a pragmatic compromise, utilizing simplified region outlines or polygonal approximations to reduce pathologists' workload. Building on this, recent research \cite{laaksonen2015refining,cheng2020self,lu2021data,Campanella,srinidhi2021deep} attempts to refine coarse annotations through deep learning methods, yet faces three main constraints: 1) heavy reliance on large-scale training data or a small subset of clean, well-curated data; 2) limited precision due to insufficient utilization of contextual information; 3) inadequate time efficiency. Although Wang et al. \cite{wang2022label} advanced this field through their LCMIL framework, a multiple instance learning approach using pseudo-positive/negative bags for coarse label refinement, their method overlooks two crucial aspects: the latent semantic patterns within and beyond the annotated regions and the cross-WSI pathological pattern correlations. 

% 引出我们的工作
To address these challenges, we propose a prototype-guided framework. First, we introduce a hierarchical prototyping approach to build a dictionary of representative pathological patterns, leveraging both intra-slide local semantics and inter-slide contextual pattern relationships to extract non-redundant and discriminative prototypes. Second, we develop a prototype-guided pseudo-labeling module that refines pseudo-labels by statistically identifying major (dominant positive) and minor (potential negative) prototype distributions with coarse annotations and aligning patches to these prototypes through similarity matching. Subsequently, the pseudo-labeled patches are utilized to train a patch classifier, where we implement a dynamic training strategy combining data sampling and re-finetuning to mitigate label noise and class imbalance. Finally, fine-grained annotations are then produced by aggregating the predictions for each patch. 

Extensive experiments are conducted on three public datasets: Camelyon16, PAIP2019, PAIP2020, covering lymph, liver, and colorectal cancers. The results demonstrate that our method significantly outperforms existing SOTA approaches in terms of precision, time efficiency and generalization capability. Our main contributions are summarized as follows: (1) We propose an innovative prototype-guided method for refining coarse annotations in WSIs, demonstrating superior performance compared to existing methods. (2) We introduce a hierarchical prototyping approach to obtain representative prototypes, leveraging both intra-slide local semantics and inter-slide contextual information. (3) We present a dynamic training strategy to effectively train a patch classifier, mitigating both label noise and class imbalance.
% (3) We present a method for effectively training a patch classifier, where the predictions are transformed into final fine-grained annotations.

\section{Method}
\subsection{Problem Formulation}\label{AA}
% When processing WSIs with coarse annotations, due to their gigapixel-level resolution, it is typical to divide them into numerous smaller patches with a fixed size ($i.e.$, 256 × 256). This can be expressed as $S^i=\left(p_j^i,y_j^i\right)_{j=1}^N$, where $i$ denotes the $i$-th WSI, and $j$ denotes the $j$-th patch. Then, based on the prior knowledge from the coarse annotations, each patch $p_j$ is assigned with a label $y_j \in\{0,1\}$.Specifically,
% patches inside the coarse annotation area are labeled as positive $(y_i=1)$, while those outside are labeled as negative $(y_i=0)$.

% However, the inherent imprecision of coarse annotations often introduces noise, leading to incorrect label assignments. For instance, negative patches may be mislabeled as positive, and vice versa. These noisy labels can mislead the model during training, reducing prediction accuracy. Thus, refining coarse annotations is essential, with the main challenge being how to effectively detect and correct erroneous labels.

% To address this challenge, we propose a prototype-guided method for refining coarse annotations by generating more accurate pseudo-labels and reducing the model's tendency to memorize noisy labels. Our approach comprises three key modules: local-to-global prototype extraction, prototype-guided pseudo-label generation, and an integrated dynamic data sampling and re-finetuning module. Fig.\ref{fig:1} depicts an overview of our proposed method, and each module will be described in detail below.

Due to the gigapixel-level resolution, WSIs are typically cropped into numerous patches with a fixed size (e.g., 256 × 256 pixels) for analysis. 
% This can be expressed as $S^i=\left(p_j^i,y_j^i\right)_{j=1}^N$, where $i$ denotes the $i$-th WSI, and $j$ denotes the $j$-th patch. 
Based on coarse annotations, 
% each patch $p_j$ can be assigned with a label $y_j \in\{0,1\}$. 
each patch \( p^j \) is assigned a label \( y^j \in \{0,1\} \). Patches within the annotated region are labeled as positive \( (y^j = 1) \), while those outside are labeled as negative \( (y^j = 0) \).
% Specifically, patches inside the coarse annotation area are labeled as positive $(y_i=1)$, while others are labeled as negative $(y_i=0)$. 
However, noise in the coarse annotation can mislead the model, resulting in performance degradation. Thus, we aim to reduce noise and obtain more accurate fine-grained annotation.

\begin{figure*}[t]
    \centering
    \includegraphics[width=\linewidth]{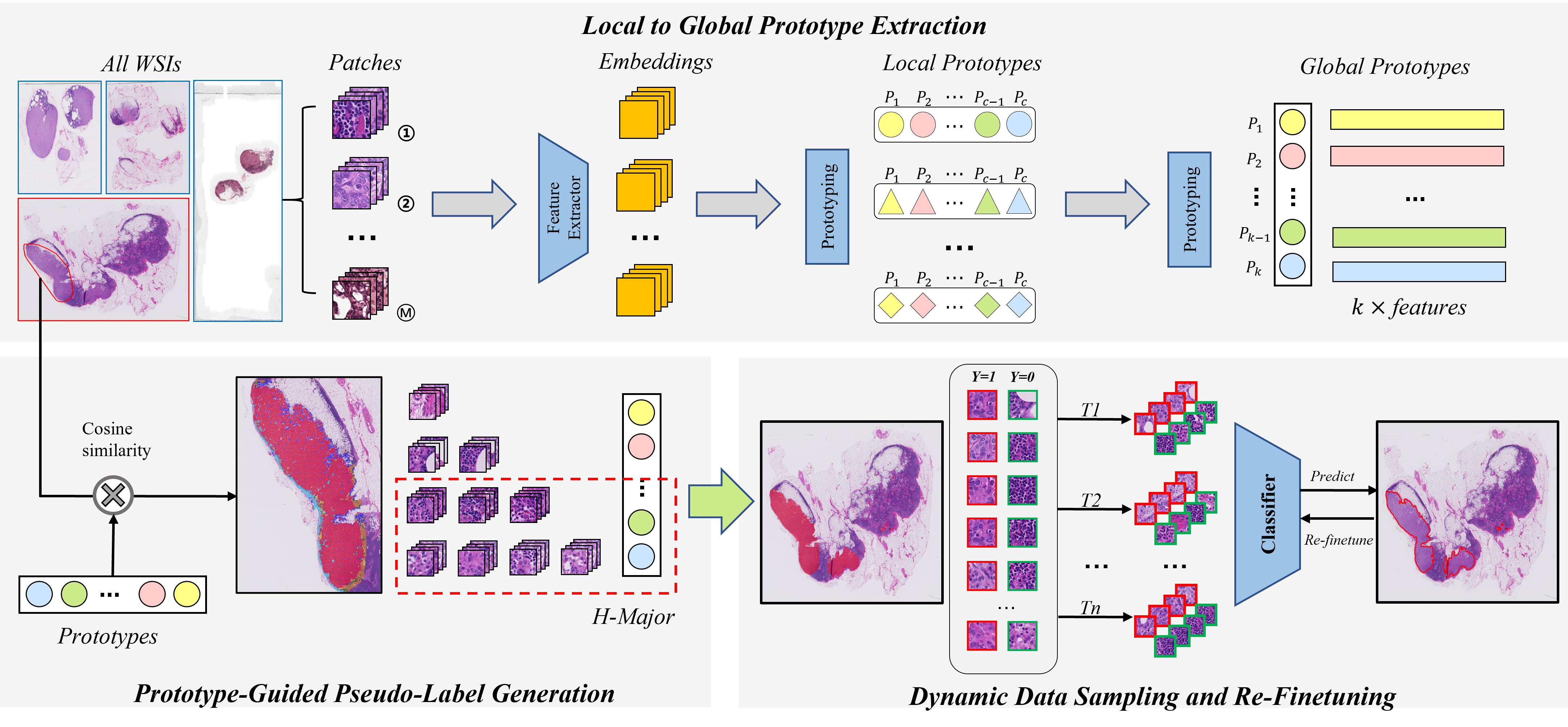}
    \caption{\textbf{Overview of our proposed prototype-guided coarse annotations refinement framework.} Local-to-global prototypes are extracted from WSI patches and utilized by the pseudo-label generation module to refine coarse annotations. Dynamic data sampling and re-finetuning module is utilized to train the patch classifier.}
    \label{fig:1}
\end{figure*}

\subsection{Overview}
We propose a prototype-guided method for refining coarse annotations, as illustrated in Fig.\ref{fig:1}. First, we introduce a local-to-global approach to extract non-redundant representative prototypes by jointly modeling intra-slide local semantics and inter-slide contextual relationships. These prototypes then guide pseudo-label refinement through statistical identification of major and minor prototypes within coarse annotations and similarity-based patch-to-prototype alignment. Subsequently, to mitigate label noise and class imbalance, we employ dynamic data sampling and re-finetuning to train a patch classifier. Finally, fine-grained annotations are produced by aggregating the predictions for each patch. Methodological details will be elaborated in the following subsections.
% We propose a prototype-guided method for refining coarse annotations, as illustrated in Fig.\ref{fig:1}. First, we introduce a hierarchical prototyping approach to
% build a dictionary of representative pathological patterns. By querying the dictionary, pseudo labels are generated for each patch and used to train a patch classifier, where a novel training strategy that includes dynamic data sampling
% and re-finetuning is introduced. Finally, fine-grained annotations are produced
% by aggregating the predictions for each patch.
\subsection{Local-to-Global Prototype Extraction}\label{AA}
WSIs typically consist of various pathological components (e.g., tumors, stroma, and lymphocytes), each exhibiting distinct characteristics \cite{song2024morphological}. Therefore, a set of key descriptors, i.e., prototypes, can effectively summarize the pathological patterns in a WSI. However, the complexity of tissue structures and the inter-WSI heterogeneity \cite{belhomme2015heterogeneity} fundamentally constrain the effectiveness of local prototypes derived from individual WSIs. These prototypes tend to exhibit two critical limitations: pattern redundancy and incomplete representation.
Therefore, we propose a local-to-global prototyping method that leverages both intra-slide local semantics and inter-slide contextual relationships to extract non-redundant and discriminative prototypes. 
% However, due to the complexity of tissue structures and the heterogeneity between different WSIs \cite{belhomme2015heterogeneity}, local prototypes within a single WSI are often limited. Local prototypes may introduce redundancy and fail to capture the core characteristics of the pathological patterns comprehensively.  Additionally, they are insufficient to represent the full diversity and commonalities of pathological components across multiple WSIs. Therefore, we propose a local-to-global prototyping method that leverages both the local semantics of images and the contextual information from datasets.

Specifically, given a WSI $S_i=\left\{p_i^j \mid p_i^j \in \mathbb{R}^{H \times W \times 3}\right\}_{j=1}^N$, where $N$ represents the patches num, we first extract patch-level embeddings $Z_i=\left\{\mathbf{z}_i^1, \ldots, \mathbf{z}_i^N\right\}$ using a visual encoder $f_{\mathrm{enc}}(\cdot)$, where $\mathbf{z}_i^j \in \mathbb{R}^d$. To distill intra-slide semantic patterns, we cluster $Z_i$ into a compact set of local prototypes $\mathcal{H}_i^{\text {local }}=\left\{\mathbf{h}_i^1, \ldots, \mathbf{h}_i^c\right\}$
with $c \ll N$. Subsequently, we further aggregate the local prototypes from $M$ WSIs, denoted as $\left\{\mathcal{H}_1^{\text {local }}, \ldots, \mathcal{H}_M^{\text {local }}\right\}$, to generate a global prototype set $\mathcal{H}^{\text {global }}=\left\{\mathbf{h}_1, \ldots, \mathbf{h}_K\right\}$ that encapsulates cross-WSI pathological patterns. For patch-to-prototype alignment, the cosine similarity between each patch $\mathbf{z}_i^j$ and global prototypes is computed as:
\begin{equation}
\operatorname{sim}\left(\mathbf{z}_i^j, \mathbf{h}_k\right)=\frac{\mathbf{z}_i^j \cdot \mathbf{h}_k}{\left\|\mathbf{z}_i^j\right\|\left\|\mathbf{h}_k\right\|}
\end{equation}
yielding a similarity vector $\mathbf{s}_i^j=\left[\operatorname{sim}\left(\mathbf{z}_i^j, \mathbf{h}_1\right), \ldots, \operatorname{sim}\left(\mathbf{z}_i^j, \mathbf{h}_K\right)\right]$. By employing this hierarchical prototyping approach, we can better capture non-redundant and discriminative prototypes.
\subsection{Prototype-Guided Pseudo-Labeling}
Due to the inherent inaccuracy of coarse annotations, two critical issues arise: 1) intra-annotation noise - potential negative regions may exist within the coarsely annotated areas; 2) extra-annotation omission - unannotated regions might contain overlooked positive patterns. Under the hypothesis that the majority of patches inside coarse annotations are positive, $i.e.$, cancerous regions, we design a prototype-guided pseudo-labeling module to address both issues.
% 由于粗标注是不准确的，所以粗标注内部可能存在一些潜在的阴性区域，同时粗标注外部也存在一些遗漏的阳性区域。我们假设粗标注内的绝大多数patches都是阳性的。然后，我们通过引入原型引导的伪标签生成模块，去排除粗标注内部的潜在阴性区域以及挖掘粗标注外部的潜在阳性区域。

Building on the hierarchical prototypes  $\mathcal{H}^{\text {global }}=\left\{\mathbf{h}_1, \ldots, \mathbf{h}_K\right\}$, for each patch $\mathbf{z}_i^j$, we assign it to the closest global prototype:
\begin{equation}
k^*=\arg \max _{1 \leq k \leq K} \operatorname{sim}\left(\mathbf{z}_i^j, \mathbf{h}_k\right)
\end{equation}
After assignment, for patches within the coarse annotations, we first quantify the frequency distribution of their assigned global prototypes in $\mathcal{H}_{\text{coarse}}^{\text{global}}=\left\{\mathbf{h}_1, \ldots, \mathbf{h}_{k^{\prime}}\right\}$. Specifically, we compute the count $n_q$, $\forall q \in\left\{1, \ldots, k^{\prime}\right\}$ of patches associated with each prototype $\mathbf{h}_q$, then rank prototypes in descending order based on their frequencies $\left\{n_1, \ldots, n_{k^{\prime}}\right\}$. The top-$m$ prototypes with the highest frequencies are selected as major-class prototypes $\mathcal{H}^{\text {major }}=\left\{\mathbf{h}_1, \ldots, \mathbf{h}_m\right\}$, which represent dominant cancer patterns. Pseudo-labels are finally assigned as:
\begin{equation}
\hat{y}_i^j= \begin{cases}1, & \max _{\mathbf{h}_m \in \mathcal{H}^{\text {major }}} \operatorname{sim}\left(\mathbf{z}_i^j, \mathbf{h}_m\right)>\theta \\ 0, & \text { otherwise }\end{cases}
\end{equation}
Through the above prototype-guided pseudo-labeling module, the information within and outside the coarse annotations is further refined.

\subsection{Dynamic Data Sampling and Re-Finetuning}
Cancerous regions typically occupy only a small portion of WSIs, resulting in an imbalance between positive and negative patches. Furthermore, although prototype-guided pseudo-labeling improves the accuracy of training samples for the patch classifier, noisy labels may still disrupt the training process. To mitigate these issues, we design a Dynamic Training module.
% Cancerous regions cover a small portion of WSIs, causing an imbalance between positive and negative samples. Despite prototype-guided pseudo-labeling improving patch classifier accuracy, noisy labels can still disrupt training. To address these issues, we designed a Dynamic Sample Training module.

Given pseudo-labeled patches from a single WSI $S=\left\{\left(p^j, \hat{y}^j\right) \mid 1 \leq j \leq N\right\}$, we randomly sample an equal number of positive and negative samples to create a balanced training batch in each training iteration. Let the positive set in the $t$-th training iteration be $P_t^{+}$and the negative sample set be $P_t^{-}$, where $P_t^{+} \subseteq\left\{(p^j, \hat{y}^j) \mid \hat{y}^j=1\right\}$, $P_t^{-} \subseteq\left\{(p^j, \hat{y}^j) \mid \hat{y}^j=0\right\}$, and $\left|P_t^{+}\right|=\left|P_t^{-}\right|=C$. Then the model is trained using the dynamically sampled data batch $P_t=P_t^{+} \cup P_t^{-}$in $t$-th iteration.
To further guide the model's focus on positive samples, we employ Focal Loss as the loss function. The loss function for the $t$-th iteration is defined as:
\begin{equation}
L_t=\frac{1}{2 C}\left(\sum_{p^j \in P_t^{+}} F L\left(f_\theta\left(p^j\right), 1\right)+\sum_{p^j \in P_t^{-}} F L\left(f_\theta\left(p^j\right), 0\right)\right)
\end{equation}
where $f_\theta$ is the classifier, and $F L$ is the Focal Loss. By balancing positive and negative samples in each iteration and emphasizing positive samples through the use of Focal Loss, the dynamic data sampling module both mitigates data imbalance and effectively reduces the impact of noisy labels on model training, thereby improving classification performance and generalization.
Since only a subset of samples is utilized in each iteration, dynamic data sampling may lead to underfitting. Therefore, we introduce Re-Finetuning, where the previously trained model $f_\theta$ is employed to predict for all patches in the WSI and update the pseudo label, where $ S^{\prime}=\left\{\left(p^j, y^{\prime \prime j}\right) \mid y^{\prime \prime j}=f_\theta\left(p^j\right), 1 \leq j \leq N\right\}$
% where $S=\left\{\left(p_j, y_j^{\prime \prime}\right)\right\}_{j=1}^N$, where $y_j^{\prime \prime}=$ $f_\theta\left(p_j\right)$.
During the re-finetuning phase, we no longer sample batches but instead train on the full dataset with the updated pseudo-labels. This allows the model to be further optimized on all available data, leading to improved accuracy and generalization.
\begin{table}[ht]
    \centering
    \caption{\textbf{Performance comparison across multiple datasets.} Best results are in bold.}
    \label{tab:1}
    \begin{adjustbox}{width=1.0\textwidth}
    \begin{tabular}{cccccccccc}
        \toprule
        \textbf{Dataset} & \textbf{Method} & \textbf{Dice} & \textbf{IoU} & \textbf{F1} & \textbf{PPV\_R} & \textbf{NPV\_R} & \textbf{TPR\_R} & \textbf{TNR\_R} & \textbf{Accuracy} \\
        \midrule
        \multirow{4}{*}{Camelyon16} & DkNN & 0.6194±0.0191 & 0.4966±0.0177 & 0.6441±0.0191 & 0.8700±0.0027 & 0.8505±0.0015 & 0.5830±0.0219 & 0.9555±0.0025 & 0.8545±0.0013 \\
        & PENCIL & 0.5480±0.0215 & 0.4697±0.0241 & 0.5601±0.0213 & 0.6992±0.0227 & 0.8663±0.0115 & 0.5243±0.0285 & 0.9684±0.0050 & 0.8614±0.0070 \\
        & LCMIL & 0.8638±0.0061 & 0.7697±0.0090 & 0.8811±0.0060 & \textbf{0.9133±0.0070} & 0.9374±0.0054 & 0.8646±0.0147 & \textbf{0.9677±0.0027} & 0.9366±0.0024 \\
        & Ours & \textbf{0.8844±0.0004} & \textbf{0.8012±0.0006} & \textbf{0.9011±0.0004} & 0.8785±0.0017 & \textbf{0.9758±0.0002} & \textbf{0.9374±0.0010} & 0.9434±0.0007 & \textbf{0.9471±0.0003} \\
        \midrule
        \multirow{4}{*}{PAIP2019} & DkNN & 0.6939±0.0008 & 0.5890±0.0009 & 0.7298±0.0008 & 0.9427±0.0086 & 0.7612±0.0005 & 0.6566±0.0009 & 0.9776±0.0020 & 0.8166±0.0005 \\
        & PENCIL & 0.5947±0.0204 & 0.5255±0.0202 & 0.6250±0.0210 & 0.7040±0.0154 & 0.7528±0.0208 & 0.5910±0.0223 & 0.9425±0.0199 & 0.8007±0.0098 \\
        & LCMIL & 0.7950±0.0066 & 0.6803±0.0070 & 0.8379±0.0064 & 0.9437±0.0187 & 0.8030±0.0098 & 0.7849±0.0211 & 0.9327±0.0148 & 0.8663±0.0030 \\
        & Ours & \textbf{0.8412±0.0002} & \textbf{0.7452±0.0002} & \textbf{0.8838±0.0001} & \textbf{0.9653±0.0002} & \textbf{0.8550±0.0001} & \textbf{0.8393±0.0001} & \textbf{0.9738±0.0002} & \textbf{0.9031±0.0001} \\
        \midrule
        \multirow{4}{*}{PAIP2020} & DkNN & 0.8264±0.0008 & 0.7109±0.0001 & 0.8045±0.1205 & 0.7884±0.0017 & 0.9498±0.0008 & 0.9530±0.0004 & 0.7919±0.0016 & 0.8763±0.0009 \\
        & PENCIL & 0.8078±0.0159 & 0.7202±0.0163 & 0.8486±0.0165 & 0.8339±0.0180 & 0.9243±0.0084 & 0.8742±0.0133 & 0.8607±0.0149 & 0.8938±0.0091 \\
        & LCMIL & 0.8425±0.0039 & 0.7402±0.0051 & 0.8468±0.0039 & \textbf{0.8977±0.0098} & 0.9036±0.0045 & 0.8900±0.0080 & \textbf{0.9342±0.0070} & \textbf{0.9099±0.0023} \\
        & Ours & \textbf{0.8515±0.0002} & \textbf{0.7493±0.0004} & \textbf{0.8872±0.0001} & 0.8444±0.0002 & \textbf{0.9410±0.0003} & \textbf{0.9443±0.0004} & 0.8611±0.0005 & \textbf0.9034±0.0001 \\
        \bottomrule
    \end{tabular}
    \end{adjustbox}
\end{table}
\begin{table}[ht]
    \centering
    \caption{\textbf{Run-Time Comparison (Minutes).}}
    \label{tab:run}
    \begin{adjustbox}{width=0.5\textwidth}
    \begin{tabular}{lccc}
    \toprule
    \textbf{Methods} & \textbf{Camelyon16} & \textbf{PAIP2019} & \textbf{PAIP2020} \\
    \midrule
    DkNN & $6.9$ & $38.9$ & $15.8$ \\
    LCMIL & $5.7$ & $18.8$ & $11.8$ \\
    Ours & $\textbf{4.8}$ & $\textbf{13.3}$ & $\textbf{7.2}$ \\
    \bottomrule
    \end{tabular}
    \end{adjustbox}
\end{table}

\section{Results}
\subsection{Experimental Setup}
\textbf{Dataset}. We utilize three publicly available datasets, Camelyon16 \cite{litjens20181399}, PAIP-2019 \cite{kim2021paip} and PAIP2020 \cite{kim2023paip} %covering a range of cancers including lymph, liver and colorectal. 
%They encompass different tissue types and exhibit morphological heterogeneity, with detailed annotations provided by pathology experts. 
and strictly adhered to the data selection and processing protocols outlined in LCMIL \cite{wang2022label}. We also incorporate the noise annotations that simulate human coarse labeling scenarios, as used in LCMIL, to further validate the effectiveness of our method in real-world applications.
\textbf{WSI Processing}. For the Camelyon16 and PAIP2020 datasets, non-overlapping patches of size 256×256 are processed at $40\times$ magnification. For the PAIP2019 dataset, the patch size is 128×128 and the magnification is $20\times$.
\textbf{Feature Extractor}. %Unlike LCMIL, which used a pre-trained VGG network on ImageNet as the feature extractor, 
We employ a pathology foundation model, CONCH \cite{lu2024avisionlanguage}, as the primary feature extractor.
\textbf{Prototyping}. We utilize K-means algorithm \cite{coates2012learning} for prototype generation.  Other prototype generation methods \cite{rymarczyk2022protomil,yang2023tpmil} can also be utilized as alternatives.
\textbf{Evaluation Metric}. %To quantitatively evaluate the refinement of coarse annotations to precise annotations, 
We use Dice, IoU, F1, Accuracy, PPV (Precision), NPV (Negative Predictive Value), TPR (Recall), and TNR (Specificity) as evaluation metrics. The final performance is reported as the average results from five repeated experiments.

\begin{figure}[h]
    \centering
    \includegraphics[width=0.85\linewidth]{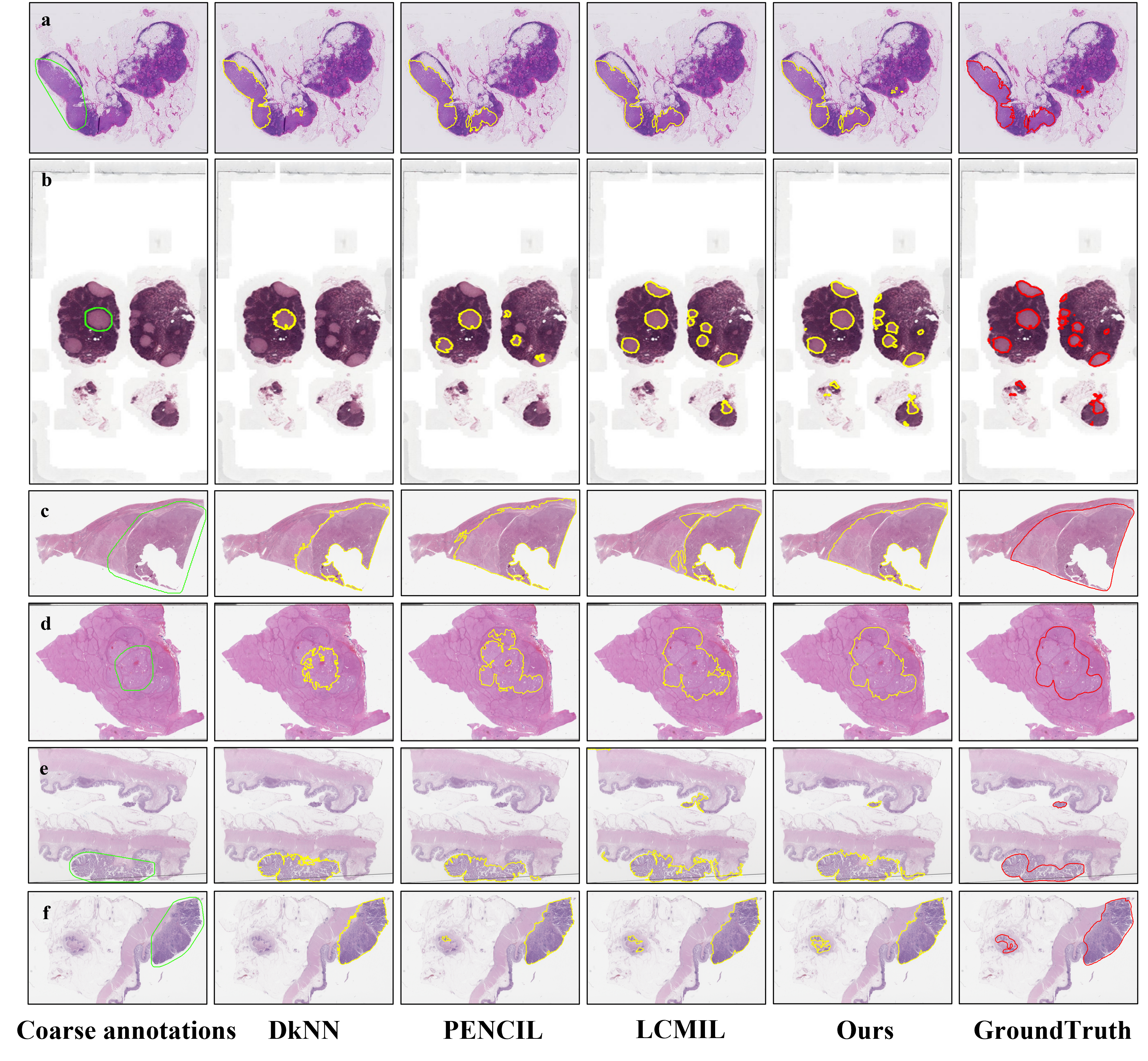}
    \caption{\textbf{Examples of coarse annotations and refinement.} (a,b), (c,d) and (e,f) are from Camelyon16, PAIP2019, and PAIP2020, respectively.}
    \label{fig:2}
\end{figure}
\subsection{Experimental Results}
\textbf{Comparative results}. We evaluate our method against the SOTA method for coarse annotation refinement in WSIs, namely LCMIL, as well as two benchmark methods for handling noisy labels in natural images, DkNN \cite{bahri2020deep} and PENCIL \cite{yi2019probabilistic}. Notably, we replace the feature extractors in these methods with the CONCH model. Table \ref{tab:1} presents the experimental results across three datasets. As shown in Tabel \ref{tab:1}, our method presents clear superiority, especially on the PAIP2019 dataset, where the Dice score shows an improvement of nearly 4.5\% comparse to LCMIL. In comparsion, DkNN and PENCIL perform less effectively. DkNN, which relies on local features through its k-nearest neighbors approach, fails to capture the broader context in complex pathology images, while PENCIL struggles with the varied noise in coarse annotations, leading to reduced performance. Furthermore, Table \ref{tab:run} compares the average time required to obtain a complete result for a single sample, highlighting our method's time efficiency relative to DkNN and LCMIL.
%achieving improvements of approximately 4.6\%, 6.5\%, and 4.6\% over LCMIL in Dice, IoU, and F1 metrics, respectively. This highlights our method's superior performance on refining coarse annotations. 
% In contrast, DkNN and PENCIL perform relatively less effectively. DkNN, which relies on local features due to its k-nearest neighbors approach, tends to miss the broader context in complex pathology images. PENCIL, although designed to handle noisy labels, faces challenges with the complex and varied noise in coarse annotations, leading to less effective performance.
% In contrast, DkNN and PENCIL perform less effectively. DkNN, focused on local features due to its k-nearest neighbors approach, misses the broader context in complex pathology images, while PENCIL struggles with varied noise in coarse annotations, reducing its performance.

\begin{figure}[h]
\centering
\includegraphics[width=1.0\columnwidth]{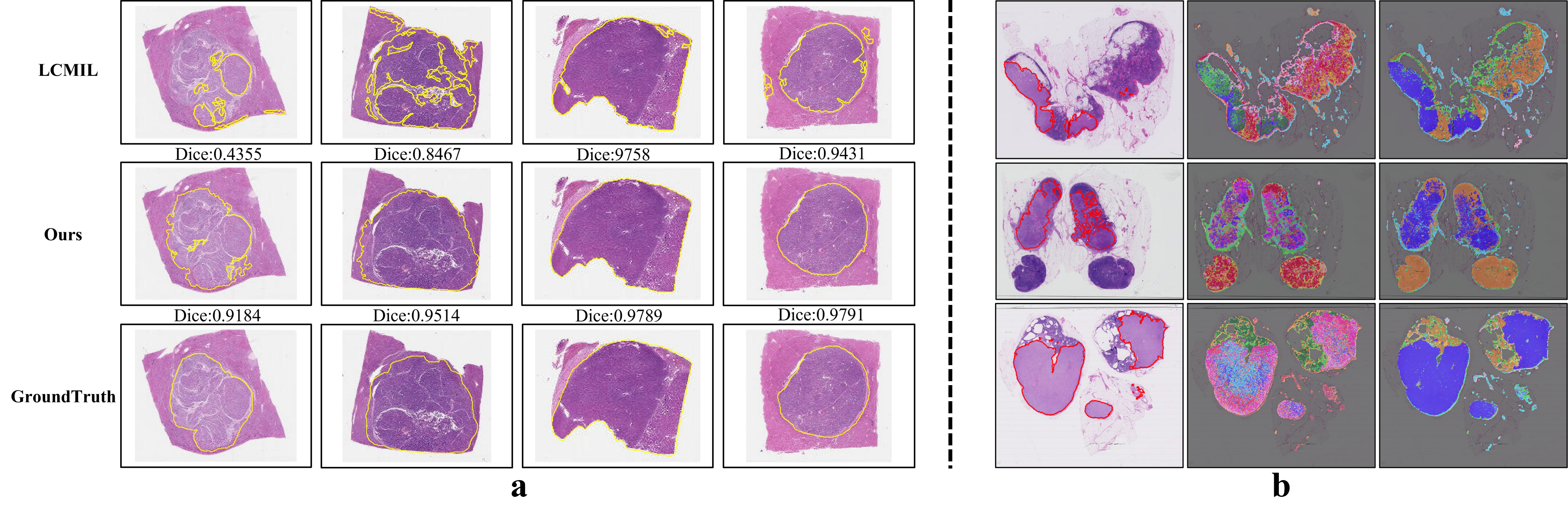} % 替换为您的图片文件名
% \caption{\textbf{Fig.3a Visualization of external validation.}}
\caption{
\textbf{(a) Visualization of external validation.}
\textbf{(b) Distribution of local prototypes and corresponding global prototypes}.The left column represents the ground truth, while the right two columns use different colors to indicate local and global prototypes, corresponding to various tissues or pathological regions.
}
\label{fig:val}
\end{figure}
% \begin{figure}
%     \centering
%     \includegraphics[width=0.5\linewidth]{图片3.png}
%     \caption{\textbf{The distribution of local prototypes and corresponding global prototypes}. The left column represents the ground truth, while the right two columns use different colors to indicate local and global prototypes, corresponding to various tissues or pathological regions.}
%     \label{fig:3}
% \end{figure}

% \begin{figure}[h]
%     \centering
%     \begin{minipage}[b]{0.6\textwidth}
%         \centering
%         \includegraphics[width=\textwidth]{external_validation.png}
%         \caption{Caption for image 1}
%         \label{fig:image1}
%     \end{minipage}
%     \begin{minipage}[b]{0.4\textwidth}
%         \centering
%         \includegraphics[width=\textwidth]{图片3.png}
%         \caption{Caption for image 2}
%         \label{fig:image2}
%     \end{minipage}
% \end{figure}
\begin{table}[ht]
    \centering
    \caption{\textbf{Ablation study on three datasets.} CA, LP, GP, DDS, RF represent  Coarse Annotations, Local Prototype, Global Prototype, Dynamic Data Sampling and Re-Finetune, respectively.}
    \label{tab:2}
    \begin{adjustbox}{width=0.8\textwidth}
    \begin{tabular}{@{}lcccccccccc@{}}
        \toprule
        Dataset & CA & LP & GP & DDS & RF & Dice & IoU & F1 & Accuracy \\ \midrule
        \multirow{5}{*}{Camelyon16} 
        & \checkmark & & & & & 0.6764 $\pm$ 0.0012 & 0.5408 $\pm$ 0.0013 & 0.6914 $\pm$ 0.0012 & 0.8634 $\pm$ 0.0007 \\
        & \checkmark & \checkmark & & & & 0.7191 $\pm$ 0.0135 & 0.5830 $\pm$ 0.0112 & 0.7355 $\pm$ 0.0135 & 0.8746 $\pm$ 0.0017 \\
        & \checkmark & \checkmark & \checkmark & & & 0.8693 $\pm$ 0.0007 & 0.7790 $\pm$ 0.0010 & 0.8872 $\pm$ 0.0008 & 0.9432 $\pm$ 0.0004 \\
        & \checkmark & \checkmark & \checkmark & \checkmark & & 0.8794 $\pm$ 0.0004 & 0.7940 $\pm$ 0.0006 & 0.8963 $\pm$ 0.0004 & 0.9437 $\pm$ 0.0002 \\
        & \checkmark & \checkmark & \checkmark & \checkmark & \checkmark & 0.8844 $\pm$ 0.0004 & 0.8012 $\pm$ 0.0006 & 0.9011 $\pm$ 0.0004 & 0.9471 $\pm$ 0.0003 \\ \midrule
        \multirow{5}{*}{PAIP2019} 
        & \checkmark & & & & & 0.5915 $\pm$ 0.0010 & 0.5015 $\pm$ 0.0008 & 0.6281 $\pm$ 0.0010 & 0.7701 $\pm$ 0.0005 \\
        & \checkmark & \checkmark & & & & 0.6964 $\pm$ 0.0033 & 0.5927 $\pm$ 0.0033 & 0.7317 $\pm$ 0.0034 & 0.8190 $\pm$ 0.0012 \\
        & \checkmark & \checkmark & \checkmark & & & 0.7901 $\pm$ 0.0011 & 0.6892 $\pm$ 0.0012 & 0.8327 $\pm$ 0.0011 & 0.8773 $\pm$ 0.0006 \\ 
        & \checkmark & \checkmark & \checkmark & \checkmark & & 0.8327 $\pm$ 0.0011 & 0.7331 $\pm$ 0.0015 & 0.8766 $\pm$ 0.0011 & 0.8973 $\pm$ 0.0007 \\ 
        & \checkmark & \checkmark & \checkmark & \checkmark & \checkmark & 0.8412 $\pm$ 0.0002 & 0.7452 $\pm$ 0.0002 & 0.8838 $\pm$ 0.0001 & 0.9031 $\pm$ 0.0001 \\ 
        \midrule
        \multirow{5}{*}{PAIP2020} 
        & \checkmark & & & & & 0.8225 $\pm$ 0.0002 & 0.7176 $\pm$ 0.0002 & 0.8635 $\pm$ 0.0001 & 0.8975 $\pm$ 0.0003 \\
        & \checkmark & \checkmark & & & & 0.7965 $\pm$ 0.0007 & 0.6773 $\pm$ 0.0008 & 0.8326 $\pm$ 0.0007 & 0.8647 $\pm$ 0.0007 \\
        & \checkmark & \checkmark & \checkmark & & & 0.8481 $\pm$ 0.0003 & 0.7467 $\pm$ 0.0005 & 0.8881 $\pm$ 0.0004 & 0.9046 $\pm$ 0.0007 \\
        & \checkmark & \checkmark & \checkmark & \checkmark & & 0.8506 $\pm$ 0.0004 & 0.7507 $\pm$ 0.0006 & 0.8892 $\pm$ 0.0004 & 0.9072 $\pm$ 0.0004 \\
        & \checkmark & \checkmark & \checkmark & \checkmark & \checkmark & 0.8515 $\pm$ 0.0002 & 0.7493 $\pm$ 0.0002 & 0.8872 $\pm$ 0.0001 & 0.9034 $\pm$ 0.0001 \\ 
        \bottomrule
    \end{tabular}
    \end{adjustbox}
\end{table}
\textbf{Visualization}. 
To provide a more intuitive comparison, we visualize several examples of the refinement performance. As shown in Figure \ref{fig:2}, our method produces more precise annotation refining, effectively identifying potential positive regions that other methods overlook. Furthermore, to explore the generalization ability of the proposed method, we test the model weights, which were trained on a random single sample, on several external liver cancer WSIs that have not been seen before, without any additional annotations. As shown in Figure \ref{fig:val}a, the model maintains excellent performance on the unseen data compared to LCMIL. These results underscore the effectiveness of the prototype information extracted by our method in capturing the characteristic features of the disease, which is particularly valuable for practical applications.
% we evaluate the model on unseen data. Specifically, we use the model checkpoint trained on a single sample and test it on several external liver cancer WSI without any annotations. As shown in Figure \ref{fig:val}, the results are quite promising. The model maintains excellent performance on previously unseen data and these results underscores the effectiveness of the prototype information extracted by our method in capturing the characteristic features of the disease.

\textbf{Ablation Study}.
%To rigorously evaluate the performance of each module in our method, we conduct comprehensive ablation experiments across three datasets. 
To evaluate the effectiveness of each module, we conduct ablation experiments.
%Dice, IoU, F1, and Accuracy were used as the primary evaluation metrics, with the results detailed in Table \ref{tab:2}. 
% As shown in Table \ref{tab:2}, the performance degrades if any component is removed or replaced. The results show that each component plays a crucial role in the proposed method.
As shown in Table \ref{tab:2}, removing or replacing any module leads to performance degradation, highlighting the importance of each module.
Notably, incorporating the global prototype module substantially enhances the model's performance compared to using only local prototypes, indicating that the global prototype effectively addresses the limitations of local prototypes by capturing cross-WSI commonalities. We also presents the distribution of local prototypes and the corresponding global prototypes for several samples. 
% As shown on Figure \ref{fig:3}, the introduction of global prototypes effectively consolidates similar prototypes, demonstrating that global prototypes achieve a representation that is both low in redundancy and high in expressiveness.
Figure \ref{fig:val}b illustrates how global prototypes consolidate similar local ones, achieving a more expressive and less redundant representation.
% To evaluate our method, we perform ablation experiments on three datasets. As shown in Table II, removing or replacing any component leads to performance degradation, highlighting the importance of each module. Notably, the global prototype module significantly boosts performance over local prototypes, addressing their limitations by capturing cross-WSIs commonalities. Figure 3 illustrates how global prototypes consolidate similar local ones, achieving a more expressive and less redundant representation.
\section{Conclusion}
In summary, we propose a prototype-guided method for refining coarse annotations in WSIs. This approach leverages prototypes to capture key patterns and features within WSIs, thereby improving the localization of cancerous regions and refining the initial coarse annotations. 
% The comprehensive evaluation demonstrates that this method is not only effective across datasets with varying degrees of cancer complexity but also exhibits excellent performance. 
Evaluation across diverse cancer datasets demonstrates the effectiveness of our method, which delivers excellent performance. Whether dealing with early-stage micro-lesions or more complex cancer morphology, our method shows stable performance, strong generalization capability and time efficiency, significantly enhancing the precision and reliability of pathological image analysis.
\bibliographystyle{splncs04}
\bibliography{ref.bib}
\end{document}